\documentclass[final,twocol]{ametsocV6.1}

\title{Improving sub-seasonal wind-speed forecasts in Europe with a non-linear model}
\thanks{This Work has been accepted to Monthly Weather Review. The AMS does not guarantee that the copy provided here is an accurate copy of the Version of Record (VoR).}

\authors{Ganglin Tian,\aff{a}\correspondingauthor{ganglin.tian@lmd.ipsl.fr}
Camille Le Coz,\aff{a}
Anastase Alexandre Charantonis,\aff{a,b}
Alexis Tantet,\aff{a}
Naveen Goutham,\aff{a,c}
and Riwal Plougonven\aff{a}
}

\affiliation{\aff{a}{LMD/IPSL, École Polytechnique, Institut Polytechnique de Paris, ENS,Université PSL, Sorbonne Université, CNRS, Palaiseau, 91120, France}\\
\aff{b}{INRIA, Paris, France}\\
\aff{c}{EDF R\&D, Palaiseau, France}
}

\abstract{Sub-seasonal wind speed forecasts provide valuable guidance for wind power system planning and operations, yet the forecast\deleted[id=rev1]{ing} skills of surface winds decrease sharply after two weeks. However, large-scale variables exhibit greater predictability on this time scale. This study explores the potential of leveraging non-linear relationships between 500 hPa geopotential height (Z500) and surface wind speed to improve \replaced[id=rev1]{sub-seasonal}{subs-seasonal} wind speed forecast\deleted[id=rev1]{ing} skills in Europe. Our proposed framework uses a Multiple Linear Regression (MLR) or a Convolutional Neural Network (CNN) to regress surface wind speed from Z500. Evaluations on ERA5 reanalysis indicate that the CNN performs better due to \replaced[id=rev1]{its}{their} non-linearity. Applying these models to sub-seasonal forecasts from the European Centre for Medium-Range Weather Forecasts, various verification metrics demonstrate the advantages of non-linearity. Yet, this is partly explained by the fact that these statistical models are under-dispersive since they explain only a fraction of the target variable variance. Introducing stochastic perturbations to represent the stochasticity of the unexplained part from the signal helps compensate for this issue. Results show that the perturbed CNN performs better than the perturbed MLR only in the first weeks, while the perturbed MLR's performance converges towards that of the perturbed CNN after two weeks. The study finds that introducing stochastic perturbations can address the issue of insufficient spread in these statistical models, with improvements from the non-linearity varying with the lead time of the forecasts.}

\begin{document}

\maketitle

\section{Introduction}\label{secIntro}
As renewable energy capacity, particularly wind energy, continues to grow, skillful sub-seasonal wind speed forecasts become important for pricing, production, transmission, and utilization of renewable energy resources. Energy producers depend on skillful sub-seasonal wind speed forecasts to plan and adjust operations of power plants, ensuring that wind turbines are active during optimal wind conditions and scheduling maintenance during low wind periods to prevent outages when demand is high, thereby enhancing efficiency in power production \citep{tawn2022subseasonal}. Grid operators require skillful sub-seasonal wind speed forecasts to maintain a stable power supply, especially in regions heavily dependent on wind power, by optimizing the integration of various energy sources, such as other renewable and fossil fuels, to meet consumer power demand \citep{cassola2012wind,chang2014literature,white2017potential}. Energy traders rely on skillful wind forecasts to estimate wind power availability and power demand, so that information on likely variations in wind speed and temperature contributes to anticipating likely changes in energy prices.

\deleted[id=rev1]{The sub-seasonal timescale, spanning three to six weeks ahead, is often referred to as a ``predictability desert" \citep{white2017potential}, serving as a challenging period between short-term weather forecasts and long-term seasonal predictions. During this challenging period, due to the chaotic nature of the atmosphere and the complexity of multi-scale interactions, initial atmospheric conditions quickly dissipate, and slowly evolving boundary conditions are yet to be established.} 
    
The predictability of surface variables like near-surface wind speed is generally lower than that of large-scale atmospheric circulation patterns on the sub-seasonal timescale. \added[id=rev1]{This is because} large-scale circulation modes\deleted[id=rev1]{, such as the El Niño-Southern Oscillation, the North Atlantic Oscillation, and the Madden-Julian Oscillation,} are closely linked with ocean-atmosphere interactions, planetary waves, and large-scale energy transfers \citep{wallace1981teleconnections,zhang2005madden}\replaced[id=rev1]{, which evolve more slowly and are less sensitive to small-scale disturbances than surface variables.}{. These dynamical processes operate over longer time scales and larger spatial scales than those of small-scale physical processes, and are relatively less affected by small-scale disturbances.}
\deleted[id=rev1]{Furthermore, large-scale variables can be generally better simulated by Numerical Weather Prediction (NWP) models than small-scale variables.} 
Existing global \replaced[id=rev1]{Numerical Weather Prediction (NWP)}{NWP} models\deleted[id=rev1]{ typically operate at spatial resolutions ranging from several kilometers to tens of kilometers} \citep{hersbach2020era5,81389,wmo2012guidelines} \added[id=rev1]{have limited ability to resolve small-scale processes that affect surface variables}. \deleted[id=rev1]{While these resolutions are suitable for modeling large-scale processes, they remain too coarse for small-scale processes.} \replaced[id=rev1]{Additionally}{Consequently}, despite significant advancements in the parameterization of small-scale phenomena, uncertainties persist within these parameterization schemes \citep{hersbach2020era5,81389}\deleted[id=rev1]{, impacting the precise modeling of surface variables}. 
\replaced[id=rev1]{T}{Furthermore, t}he initialization of variables \replaced[id=rev1]{also presents a } {presents an additional }challenge: while large-scale atmospheric fields can be initialized accurately using direct satellite and radiosonde observations, the initialization of surface variables is often limited by the inconsistent quality and incomplete global coverage of in situ surface data \citep{hersbach2020era5,bauer2015quiet,81389}. 
\deleted[id=rev1]{As a result, surface variables such as near-surface temperature, wind speed, and precipitation are characterized by lower predictability, stemming from their sensitivity to local conditions and small-scale processes \protect\citep{jimenez2010surface,pielke2013mesoscale, lorenz1969predictability}.}

Leveraging the predictable information in large-scale variables can significantly improve the skill of sub-seasonal forecasts for surface variables \citep{mariotti2020windows,vigaud2017multimodel,bueler2020stratospheric}. In-depth analysis of the relationship between large-scale phenomena and surface wind speed has revealed both linear and non-linear statistics. \citet{alonzo2017modelling} and \citet{goutham2023statistical} have developed and refined methods such as Principal Component Analysis (PCA) and Redundancy Analysis (RDA) to capture these complex interactions between the geopotential height at
500 hPa (Z500) and surface wind speed in a linear way, to improve the forecast\deleted[id=rev1]{ing} skill of wind speed. However, the relationships between large-scale variables and surface variables are often complex, containing both linear and non-linear components\citep{salameh2009statistical,wilby1997downscaling,chen2011uncertainty,maraun2010precipitation,sachindra2018statistical}. 
Neural networks effectively handle non-linear relationships in high-dimensional meteorological data, capturing complex interdependencies between large-scale and surface variables without requiring specific data distribution assumptions \citep{maraun2010precipitation,sachindra2018statistical,pan2019improving,rodrigues2018deepdownscale}. This advantage is particularly valuable for meteorological applications, where data often follows complex, non-normal distributions. However, the primary focus has been on surface temperature and precipitation \citep{wilby1997downscaling,sachindra2018statistical,pan2019improving,rodrigues2018deepdownscale}, with less attention given to surface wind speed. Capturing these complex relationships accurately is a potential approach to improving wind speed forecast\deleted[id=rev1]{ing} skill. Yet, even with accurate representation of these relationships, the intrinsic uncertainty in atmospheric conditions on the sub-seasonal timescale poses ongoing challenges that need to be addressed. 
\replaced[id=rev1]{To address forecast uncertainty,}{The uncertainty in sub-seasonal forecasting is primarily from the chaotic nature of the atmosphere, the uncertainty in initial conditions, and the limitations of NWP models. Atmospheric systems, being complex and non-linear, are inherently sensitive to initial conditions \protect\citep{lorenz1969predictability}. Even minor errors in these conditions can rapidly amplify during forecasting time scales, leading to significant biases. Compounding this issue, NWP models, despite significant advancements over recent decades, still face challenges such as the parameterization of physical processes, constraints on resolution, and computational capacity \protect\citep{hersbach2020era5,bauer2015quiet,81389}. To quantify and address this uncertainty,} 
meteorologists employ ensemble forecasting techniques. \replaced[id=rev1]{Sub-seasonal forecasting systems often suffer from under-dispersion, where ensemble spread inadequately captures the actual forecast uncertainty \protect\citep{robertson2015improving,bi2022pangu,kurth2023fourcastnet,chen2023fuxi,orth2014using}. \protect\citet{zhu2018toward} demonstrated that improving uncertainty quantification and representation can significantly enhance forecast skill.}{ This approach involves running multiple forecasts with slightly different initial conditions and model parameterizations, with ensemble spread reflects the statistical distribution of forecast errors. However, many sub-seasonal forecasting systems often exhibit insufficient ensemble spread (under-dispersion), potentially underestimating actual forecast uncertainty \protect\citep{robertson2015improving,bi2022pangu,kurth2023fourcastnet,chen2023fuxi,orth2014using}.This under-dispersion can lead to overly confident forecasts that fail to adequately reflect the range of possible atmospheric states. However, \protect\citet{zhu2018toward} has shown the potential to improve the forecasting skill by better uncertainty quantification and representation.}

While substantial progress has been made in medium-range forecast up to two weeks \citep{price2023gencast,rasp2023weatherbench,lam2022graphcast} and seasonal forecast up to 13 months \citep{alonzo2017modelling}, \replaced[id=rev1]{yet the sub-seasonal timescale, spanning three to six weeks ahead, presents distinct challenges as both initial atmospheric conditions and slowly evolving boundary forcings need to be considered}{the sub-seasonal timescale has received comparatively less attention}. Advancing sub-seasonal wind speed forecast skill requires both a deep understanding of the complex relationships between large-scale variables and surface wind speed, and the accurate quantification of ensemble uncertainty. To improve sub-seasonal wind speed forecasts, this study aims to explore both the linear and non-linear relationships between large-scale atmospheric variables and surface wind speed, while also addressing the under-dispersion of sub-seasonal forecasting ensembles. Our research is guided by two key questions:

\begin{itemize}
 \item Can we leverage non-linear regression relationships between large-scale variables and surface wind speed to improve sub-seasonal forecast\deleted[id=rev1]{ing} skill over Europe, surpassing the performance of linear models?
 \item How can we effectively represent ensemble dispersion when applying the regression relationships to dynamical forecasts on sub-seasonal scales?
\end{itemize}

This study is structured as follows: section \ref{secMethodology} introduces our methodology, starting with our framework, followed by a description of our linear and non-linear regression models and their training strategies. In section \ref{secCase}, we delve into the specifics of the cases studied, including the rationales behind the choice of input and target variables and data sets. Section \ref{secRegression} presents the performance of our regression models on a historical dataset. Subsequently, in section \ref{secForecasting}, we conduct probabilistic ensembles, applying the trained regression models to each ensemble member to assess whether the chosen non-linear model can outperform the linear model in improving forecast\deleted[id=rev1]{ing} skill for wind speed on the sub-seasonal timescale. Finally, section \ref{secConclusion} offers a summary.

\section{Methodology}\label{secMethodology}
\subsection{Framework}
\begin{figure}[t]
  \centering
  \noindent\includegraphics[width=19pc,angle=0]{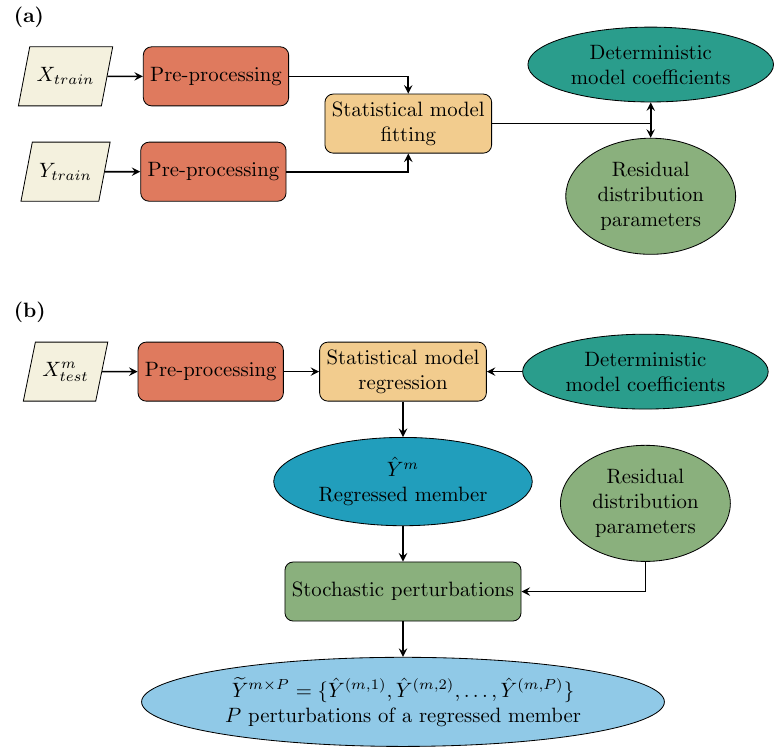}\\
  \caption{Our framework for improving surface-scale variable forecast\deleted[id=rev1]{ing} skill comprises two main stages: \textbf{(a)} Training stage: Both deterministic inputs $X$ and deterministic targets $Y$ undergo identical data pre-processing (appendix, section \ref{appendix:Preprocessing}) before being used to train a regression model. The outputs of the training stage comprises the optimal model coefficients of the statistical model (Sec~\ref{subSecModel}) and the parameters of the residual distribution (Sec~\ref{subSecStochastic}). \textbf{(b)} Ensemble forecasting stage: The $m$-th member $X^m$ of an input ensemble, after pre-processing, is fed into the trained model to regress the corresponding regressed ensemble member $\hat{Y}^m$. Then we randomly sample $P$ times from the residual distribution from the training stage to perturb the regressed member $\hat{Y}^m$, in order to obtain $P$ perturbed members $\widetilde{Y}^{m \times P}$ for this single regressed member. }
   %
  
    \label{fig:workflow}
\end{figure}

We introduce a two-stage framework (as in figure \ref{fig:workflow}) to improve the sub-seasonal forecast\deleted[id=rev1]{ing} skill of a surface variable, incorporating a) training and validation of a regression model on historical data (reanalysis, see section \ref{secCase}\ref{subsecData}) and \replaced[id=rev1]{b) applying the regression model to ensembles}{b) ensemble forecasting applying the regression model}. In the training and validation phase, as in figure \ref{fig:workflow}a, we aim to establish a mapping between a large-scale variable $X$ and a surface variable $Y$ using the statistical information from historical deterministic data. On the training dataset, the inputs $X_{train}$ and the targets $Y_{train}$ are pre-processed using identical data pre-processing procedures (see appendix, section \ref{appendix:Preprocessing}). Subsequently, optimal parameters for the regression model are obtained by minimizing a loss between the estimated outputs $\hat{Y}_{train}$ and the targets $Y_{train}$ (see section \ref{secMethodology}\ref{subSecModel}). Additionally, the differences between $\hat{Y}_{train}$ and $Y_{train}$ are used to derive the distribution parameters of their residuals (see section \ref{secMethodology}\ref{subSecStochastic}).

In the ensemble forecasting stage in figure \ref{fig:workflow}b, in order to produce $M$-member forecasting ensembles $Y^M$ of the target variable, the trained regression model is applied to $M$-member forecasting ensembles $X^M$ of the input variable. The $m$-th member $X^m$ of the input variable independently feeds into the regression model to yield \replaced[id=rev1]{the corresponding regressed}{a corresponding the regressed} $m$-th member $\hat{Y}^m$. The outputs, $M$-member ensembles produced by the regression model, are termed the regressed ensembles $\hat{Y}^M$. 
However, the regression model captures only a portion of the variance of the target, as indicated by Mean Square Error (MSE) (detailed further in section \ref{secRegression}). Consequently, when the model is applied to an ensemble dataset, the outputs exhibit reduced variance relative to the expected variance of the target. This leads to a systematic  under-estimation of dispersion and variability of the target members. 
To maintain the total variance of the target and to attribute skill improvements more accurately to the representation of the predictable components of the signal, rather than to the variance of the unpredictable components, we employ a perturbed version of the model. For the $m$-th regressed output member $\hat{Y}^m$, $P$ perturbations are randomly drawn from the residual distribution to perturb the $\hat{Y}^m$. This process produces $P$ perturbed members for $m$-th member, denoted as $\widetilde{Y}^{m\times P}$ (refer to section \ref{secMethodology}\ref{subSecStochastic}).

In this study, the results obtained through the regression models are termed ``statistical," while those based solely on numerical weather prediction models are termed ``dynamical." To differentiate between model outputs when using reanalysis versus ensemble members as inputs, the term ``prediction" is avoided. The outputs based on reanalysis are referred to as ``regression outputs," and the outputs deriving from ensemble members are called ``regressed ensembles". We use superscripts to differentiate between deterministic and probabilistic data. $X$ and $Y$ denote deterministic input and target dataset, respectively, while $X^M$ and $Y^M$ represent the input and target datasets comprised of M members, with the $m$-th member expressed as $X^m$ and $Y^m$. The notation `` $\hat{}$ " is employed to indicate values that are regressed. The ensembles and the models in which perturbations have been included are indicated with a tilde `` $\widetilde{}$ ". Lastly, all physical variables (such as wind speed and geopotential height) are represented as fields on latitude-longitude grids.

\subsection{\replaced[id=auth]{Model architectures}{Models architecture}}\label{subSecModel}

As previously mentioned, \citet{alonzo2017modelling} and \citet{goutham2023statistical} used the linear relationships based on PCA and RDA to regress surface wind speed from Z500. This study explores whether the non-linearity between large-scale variables and surface wind speed contributes to further improving the ensemble forecast\deleted[id=rev1]{ing} skill of surface wind speed, compared to linear models, rather than determining the most skillful model structure. Therefore, we opted for a simple Multiple Linear Regression model (MLR) to capture the linear relationship between inputs and a target. In our case, we developed the MLR using multiple grid points of Z500 over an input domain (more details can be found in section \ref{secCase}\ref{subCase}) to regress U100 at a specific grid point.

Acknowledging the inherent spatial patterns in Z500 \citep{goutham2023statistical}, the sea-land distribution differences in U100, and the non-linearity from Z500 to U100 \citep{salameh2009statistical}, we adopted a Convolutional Neural Network (CNN) model to capture this non-linearity. CNNs, through multiple layers of convolution operations and non-linear activation functions, effectively learn spatio-temporal patterns and non-linear relationships within data. Building on this, we adopted SmaAt-UNet \citep{trebing2021smaat}, a variant of UNet \citep{ronneberger2015u}. Compared to the standard UNet, which has been widely applied in the field of meteorology \citep{nguyen2022convolutional,bouget2021fusion}, SmaAt-UNet integrates attention modules and depthwise separable convolutions, significantly reducing the number of parameters while maintaining model performance. 
The breadth of this model family is related to CNN architectures, and it is uncertain whether more intricate non-linear relationships could yield additional explanatory power. However, our experiments indicate that within the constraints of the available training set, augmenting complexity through the integration of additional layers and channels does not enhance the MSE of the regressed deterministic U100. \replaced[id=auth]{The architectures of MLR and CNN, and their training configurations are detailed in appendix, section \ref{appendix:models}.}{These model are implemented on Pytorch \citep{paszke2019pytorch}, and are available in our GitHub repository.}

\subsection{Stochastic forecasts}\label{subSecStochastic}

The reduction in variance associated with the deterministic regression models significantly affects the assessment of probabilistic ensembles. The skill of these ensembles, as quantified by Continuous Ranked Probability Skill Score (CRPS; see section \ref{secCase}\ref{subSecScores}), is intrinsically linked to their reliability and includes their capacity to capture the variance (as evaluated by Spread Skill Ratio (SSR; see section \ref{secCase}\ref{subSecScores})). This lost variance from the deterministic regression does not reflect a decrease in forecast uncertainty; rather, it is of a purely statistical nature during the training stage in figure \ref{fig:workflow}a. Consequently, one approach to preserve the variance that represents dynamical uncertainty is the introduction of stochastic perturbations. This stochastic perturbation approach should account for the portion of the variance that the deterministic regression model fails to explain. In other machine-learning-based ensemble forecasting studies \citep{bi2022pangu,kurth2023fourcastnet,chen2023fuxi,orth2014using}, their statistical models did not account for these systematic errors, potentially leading to the under-dispersion of ensembles. We represent the unexpressed fluctuations as residuals, which are the differences between $Y_{train}$ and $\hat{Y}_{train}$. These residuals, representing the uncertainty in the model's fit to a deterministic training dataset as systematic errors, are modeled with a Gaussian distribution to estimate their mean and variance as stochastic perturbations in figure \ref{fig:workflow}a. As introduced at the beginning of this section, to express this uncertainty during the ensemble forecasting phase in figure \ref{fig:workflow}b, for each regressed member $\hat{Y}^{m}$, $P$ perturbations are randomly sampled from this distribution and then added to $\hat{Y}^{m}$ to obtain $P$ perturbed members $\widetilde{Y}^{m \times P} = \{\hat{Y}^{(m,1)},\ldots,\hat{Y}^{(m,P)} \}$. Our experiments (not shown) indicate that the ensemble skill converges when $P=20$. 

\subsection{Scores and significance test}\label{subSecScores}
We evaluate the regressed outputs $\hat{Y}$ and the ensembles $\hat{Y}^M$ and $\widetilde{Y}^M$ against their corresponding reference $Y$. All scores are evaluated grid-point by grid-point separately and later spatially averaged with cosine-latitude weights over a domain of interest. The same approach is followed with respect to the lead times. During the regression validation phase in figure \ref{fig:workflow}a, we evaluate the ability to capture the relationship between Z500 and U100 on reanalysis, quantified by the MSE between the ground truth $Y_{validation}$ and the regressed outputs $\hat{Y}_{validation}$ on the validation set of a historical reanalysis. Subsequently, in the ensemble forecasting phase in figure \ref{fig:workflow}b, we measure the forecast\deleted[id=rev1]{ing} skill of the regressed ensembles $\hat{Y}_{test}^M$ and the perturbed ensembles $\widetilde{Y}_{test}^{M \times P}$ at each independent lead time by MSE of ensemble mean and CRPS \citep{wilks_statistical_2019, matheson1976scoring, zamo2018estimation}. Additionally, ensemble reliability is quantified by SSR \citep{rasp2023weatherbench}. MSE quantifies the errors in deterministic forecasts, while CRPS assesses the accuracy of probabilistic forecasts by evaluating the alignment between ensemble distributions and reanalysis, while SSR measures the relative dispersion of ensembles. The formulas for these metrics are detailed in appendix, section \ref{appendix:score}. 

To assess the significance of the relative improvements of a statistical model $model_s$ exhibited over a benchmark $model_b$, for a given score, we employ a bootstrap technique similar to that described by  \citet{goddard2013verification}. This involves randomly selecting identical samples of ensembles from $model_s$ and $model_b$, with replacement, to generate a new subset of samples, with as many samples as in the original samples. These samples are then used to calculate the aforementioned verification scores. We compute the relative difference in scores $\Delta_r {Score} = \frac{Score(model_s) - Score(model_b)}{Score(model_b)} \times 100 (\%)$ between the two models and repeat this process 1000 times. For negatively oriented scores, such as MSE and CRPS, the proportion of $\Delta_r Score<0$ serves as the $p$-value. Conversely, for positively oriented scores, the proportion of $\Delta_r Score >0$ is used as the $p$-value. If the $p$-value is less than the predetermined significance level $\alpha$, the $model_s$ is considered significantly better than $model_b$ ($p < \alpha$).

\section{Case and Data}\label{secCase}
\subsection{Case study}\label{subCase}
We apply the above framework to a specific case study. Here, the input variable is the geopotential height at 500 hPa, Z500, over the Europe-Atlantic domain (20°–80°N, 120°W–40°E). The output variable is the wind speed at 100-meter, U100, over the European region (34°–74°N, 13°W–40°E). These regions of interest are shown in figure \ref{fig:domain_large_orography}.

\subsubsection{Target variable: } 
To study wind speed at hub height on a sub-seasonal timescale across the Europe domain in figure \ref{fig:domain_large_orography}, we focus on U100, as in  \citet{goutham2023statistical}, obtained from the zonal (u) and meridional (v) components of 100-meter wind using the formula: $U100 = \sqrt{(u^2 + v^2)}$.

\subsubsection{Input variable:}
\begin{figure}[t]
  \noindent
  \centering
  \includegraphics[width=19pc,angle=0]{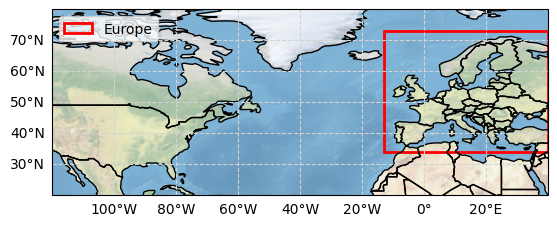}\\
  \caption{The two domains of interest: the Europe-Atlantic domain (20°–80°N, 120°W–40°E) for the input variable Z500 and the Europe domain (34°–74°N, 13°W–40°E) for the target variable U100.}\label{fig:domain_large_orography}
\end{figure}

The ensemble forecast\deleted[id=rev1]{ing} skill of wind speed declines rapidly with increasing lead time; the U100 ensembles are skillful for less than about ten days \citep{buizza2015forecast,goutham2022skillful}. When selecting input variables, we aim to choose those that remain skillful within a two to six-week time scale. Z500, which reflects the large-scale atmospheric circulations in the mid-troposphere and is commonly associated with atmospheric fluctuations and 500 hPa geostrophic winds, is extensively used in the studies examining the regression relationship between large-scale variables and wind speed \citep{alonzo2017modelling, goutham2023statistical,liu2023north}. Moreover, the Z500 has higher predictability on the sub-seasonal timescale than the 10-meter wind  \citep{toth2019weather}. \citet{buizza2015forecast} demonstrated that Z500 remains more skillful than climatology up to approximately lead times of 22 days. In this study, Z500 is cropped into the Europe-Atlantic domain as depicted in figure \ref{fig:domain_large_orography}. This specific domain is selected because the target domain for U100 is Europe, where downstream dependency of forecast errors typically occurs at mid-latitudes \citep{simmons1979downstream}. Consequently, the domain extends further westward relative to the target domain than eastward to capture its large-scale circulations.

\subsubsection{The season of interest: }
We validate our methodology during winter. This decision is based on the fact that Z500 exhibits the most significant anomalies during winter \citep{buizza2015forecast}. Additionally, the correlations between atmospheric circulations and surface variables vary seasonally and are strongest in this season \citep{laurila2021climatology}. Moreover, the higher frequency of low-pressure systems during winter leads to large wind speed variability, posing challenges for the energy sector and necessitating skillful estimations of U100. 

\subsection{Data}\label{subsecData}
\subsubsection{Historical deterministic dataset: }
We employ the ECMWF Reanalysis v5 (ERA5) high-resolution reanalysis dataset as our reference. Reanalysis combines historical meteorological observations with NWP models to generate physically consistent descriptions of past atmospheric states, widely serving as a reference in ensemble forecasting \citep{hersbach2018era5pressure,hersbach2018era5}. We source Z500, the u and v components of 100-meter wind reanalysis data spanning from December 1979 to March 2022, with a spatial resolution of 2.7 degree and temporal resolution of 6 hours from Climate Data Store (CDS). We then downsampled the Z500 and U100 reanalysis data to a weekly temporal resolution during the pre-processing step (discussed further in subsequent paragraphs). 

\subsubsection{Ensembles dataset: }
We use the ECMWF extended-range hindcasts \citep{ECMWF_Technical_Memoranda} as the dataset for ensemble forecasting. Hindcasts, generated through NWP models for retrospective periods, help calibrate and assess forecasting models' performance. The ECMWF hindcasts are initialized from ERA5 reanalysis data and generated with a control member and ten perturbed members under slightly different initial atmospheric and oceanic conditions with stochastic parameterizations. In this study, we use only these ten perturbed members, which are expected to sample part of the errors that grow from uncertain initial conditions and part of the uncertainty associated with model errors. 

Besides hindcasts, ECMWF also provides forecast ensembles, which are estimates of the atmospheric state over specific future periods based on current observations and NWP models. Unlike hindcasts, forecasts are initialized from analysis and consist of 50 perturbed members and a control member. ECMWF extended-range forecasts are available from 2015 and thus cover a shorter period than the hindcasts. This study primarily uses hindcasts for verification owing to their extended temporal coverage, encompassing a range of meteorological phenomena, including extreme weather events and seasonal variations, thereby facilitating a comprehensive evaluation of model performance. Unless stated, the dynamical ensembles mentioned hereafter refer to hindcasts, while the results for forecasts are also provided in appendix, section \ref{appendix:res_fc}. The forecasts and hindcasts of Z500 and U100 were obtained from ECMWF via Meteorological Archival and Retrieval System (MARS). We downloaded the hindcasts initialized in December, January, and February (DJF) from 1995 to 2021 and the forecasts in DJFs from 2015 to 2021 from MARS, covering 128 initializations for forecast ensembles and 2560 initializations for hindcast ensembles. The Integrated Forecasting System (IFS) of ECMWF has undergone several updates during these periods; however, the statistical differences between these various versions are expected to be marginal \citep{goutham2022skillful}. In this study, we employ hindcasts only up to March \replaced[id=rev1]{2022}{2023}, due to data availability. 

\subsubsection{Spatio-temporal resolution: }
Predictability depends on spatial and temporal scales \citep{vitart2014evolution}. Small-scale features fluctuate more frequently than large-scale features, making them less predictable over sub-seasonal timescales \citep{vitart2019introduction}. A simple method to reduce unpredictable noise is averaging data over time and space \citep{buizza2015forecast}. Here, bilinear interpolation is applied to achieve a desired resolution of $2.7\times2.7$ degree (approximately 300 km) for both the U100 and Z500 ensembles, leading to $(22\times59)$ grid points for Z500 over Europe-Atlantic domain and $(14\times19)$ grid points for U100 over Europe domain. These ensembles are averaged weekly to focus on the potentially predictable parts of the signal. The reanalysis are also aggregated at the same spatio-temporal resolution to maintain coherence between different datasets.

\begin{figure*}[t]
  \noindent
  \centering
  \includegraphics[width=39pc,angle=0]{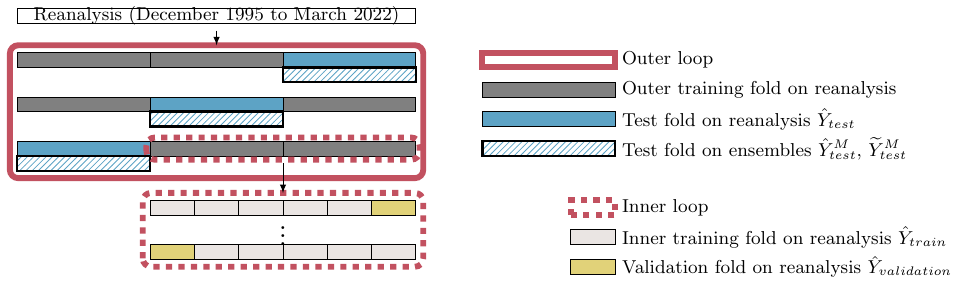}
  \caption{Nested cross-validation process consisting in two layers: the outer layer enclosed in solid red lines and the inner layer in dashed red lines. In the outer layer, each rectangle represents a period of 9 years. The gray rectangles indicate the training reanalysis folds, the blue rectangles signify the test reanalysis folds, and the rectangles with blue hatching depict the test ensembles' folds. In the inner layer, each rectangle spans 3 years, with gray rectangles representing the training reanalysis folds and light yellow rectangles indicating the validation reanalysis folds.}\label{fig:nestedCV}
\end{figure*}

\subsubsection{Nested cross-validation: } 
Nested cross-validation is extensively employed for cross-validation on small datasets to prevent overfitting. We reserve 15 years of reanalysis as climatology for data pre-processing (in appendix, section \ref{appendix:Preprocessing}), thus limiting the size of data available for neural network training from December 1995 to March 2022. This relatively short period may lead to overfit the neural networks; therefore, we use nested cross-validation for such small datasets to optimize model selection and hyperparameter configurations while ensuring the generalization capability.
Our nested cross-validation comprises outer and inner cross-validation layers, as illustrated in figure \ref{fig:nestedCV}. For the outer layers, we divide the 27 years of reanalysis data, from December 1995 to March 2022, into three 9-year folds. For each outer split, one fold is reserved as a test set (illustrated in blue in figure \ref{fig:nestedCV}) to verify the performance of deterministic outputs $\hat{Y}_{test}$ obtained by regressing using the Z500 reanalysis as input. The remaining two folds are used as the training set (shown in gray in figure \ref{fig:nestedCV}), further split by the inner split for inner training and validation. Therefore, within each outer fold's 18-year training dataset, we further divide it into six 3-year inner folds for subsequent cross-validation. Here, one fold is kept as a validation set (depicted in light yellow in figure \ref{fig:nestedCV}), and the other five serve as the training set (again shown in gray in figure \ref{fig:nestedCV}). During the inner cross-validation, various hyperparameter combinations are trained and validated to identify the optimal configuration. The selected hyperparameter combination is then evaluated on the corresponding outer fold's reanalysis test set and the paired hindcasts test fold. To prevent data leakage, we partition the 27 years of hindcast data $\hat{Y}_{test}^M$ based on their initialization dates to align with the temporal segments of each outer test fold (represented by the hatched blue rectangles in figure \ref{fig:nestedCV}). We then average the skills from these three test folds to represent the final skill of our methodology.

\subsection{Skill of the dynamical ensembles for Z500 and U100}\label{subsecZ500SkilfulThan100uv}

\begin{figure*}[t]
  \noindent
  \centering
  \includegraphics[width=39pc,angle=0]{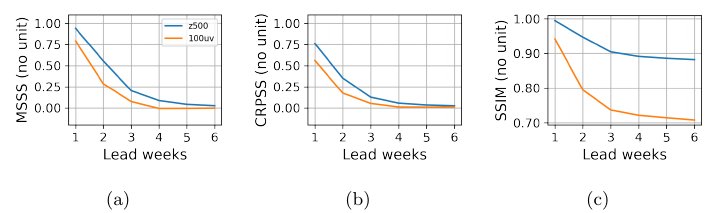}
  \caption{ The spatially averaged MSSS, CRPSS and the SSIM are displayed as a function of lead week averaged across the Europe-Atlantic domain for Z500 and the Europe domain for U100. 
  }\label{fig:check_hypo_1}
\end{figure*}

\citet{toth2019weather} have noted that large-scale/low-frequency variables possess a longer forecasting horizon compared to small-scale/high-frequency variables, consistently with expectations from geophysical fluid dynamics. Here, we revisit this by quantifying how much more skillful is Z500 relative to U100 in the case considered. Since Z500 and U100 represent different physical variables, the predictability of these variables is compared using relative skill scores \citep{wilks_statistical_2019}, such as Mean Squared Skill Score (MSSS) in figure \ref{fig:check_hypo_1}a and Continuous Ranked Probability Skill Score (CRPSS) in figure \ref{fig:check_hypo_1}b, for Z500 and U100 with respect to their respective 15-year rolling climatology reference. The formulas for MSSS and CRPSS are detailed in appendix, section \ref{score_msss_crpss}. MSSS assesses the deterministic skill of the ensemble mean for Z500 and U100 hindcasts, which decreases over time. In figure \ref{fig:check_hypo_1}a, U100 converges by lead week 4, while Z500 continues to exhibit a declining trend up to lead week 6. During this period, Z500 demonstrates significantly higher MSSS than U100, indicating that, the ensemble mean of Z500 is more predictable than that of U100. CRPSS evaluates the probabilistic skill of the ensemble members for Z500 and U100 dynamical hindcasts, which also decrease over time. In figure \ref{fig:check_hypo_1}b, up to lead week 6, Z500 consistently shows higher CRPSS than U100, suggesting that the distribution of the Z500 hindcasts aligns more closely with the corresponding reanalysis. These findings from the MSSS and the CRPSS reflect the inherent challenges in forecasting U100, which is less predictable than Z500. However, these scores do not account for spatial correlations for each variable. 

We employ Structural Similarity Index (SSIM) as a relative metric to assess the spatial similarity between the dynamical ensemble members of Z500 and U100 and their corresponding reanalysis. SSIM, which accounts for variations in luminance, contrast, and local structure of two images, measures the spatial mean similarity, spatial variance similarity, and normalized spatial covariance, as shown in equation \eqref{eq:ssim} in appendix, section \ref{appendix:score}. Predominantly utilized in computer vision \citep{wang2004image}, SSIM values range from 0 to 1, where 1 denotes perfect identity between images. SSIM quantifies the spatial covariance between the hindcasts and their respective reanalysis, facilitating the understanding of the spatial variations in the atmospheric variables across various lead times. Figure \ref{fig:check_hypo_1}c presents SSIM for Z500 and U100 across lead times. The SSIM values for both Z500 and U100 exhibit a decreasing trend, with a sharp decline after lead week 2, and a convergence after lead week 4. This trend implies a progressive decrease in the dynamical forecast\deleted[id=rev1]{ing} skill of the ECMWF hindcasts over time. Up to 6 weeks, Z500 always exhibits a higher SSIM compared to U100, indicating greater predictability for Z500.

To gain a deeper understanding of the impacts of the spatial mean, the spatial variance and the spatial covariance on the Z500 and U100 hindcasts, we delineate the three components of SSIM in appendix, section \ref{appendix:hypothese}. As evident from figure \ref{fig:check_hypo_appendix}a in appendix, the structure component exhibits a similar declining trend to the SSIM and serves as the primary factor on the SSIM decreasing trend. Whereas the variations in the luminance component and the contrast component remain relatively minor (as in figures \ref{fig:check_hypo_appendix}b and \ref{fig:check_hypo_appendix}c in appendix), indicating a modest change in the dynamical spatial-averaged bias and the dynamical spatial variability, respectively. In other words, these dynamical hindcasts do not become more blurred, and the primary loss in predictability stems from changes in the spatial structure of dynamical hindcasts, particularly the misrepresentation of features, such as local pressure lows and associated fronts. When comparing the relative skills of Z500 to U100, it is evident that Z500 consistently exhibits higher SSIM and structure. 
Due to dynamical factors, it is anticipated that \replaced[id=rev1]{Z500}{pressure} is a larger-scale and more slowly evolving field compared to wind. Furthermore, a mid-tropospheric field, characterized by its large scale, tends to be more predictable than a surface field.\deleted[id=rev1]{ Note that these relative scores are only used in this section to compare the skills of the different variables, and confirm quantitatively the greater predictability of Z500. This is consistent with expectation from geophysical fluid dynamics: Through hydrostatic balance, pressure represents the mass of the atmosphere above the level of interest. Logically this can only vary slowly and on large spatial scales. Wind, in contrast, is to a first approximation close to spatial derivatives of pressure (geostrophic balance) and hence includes small scales. } That Z500 exhibits the higher MSSS, CRPSS and SSIM values compared to U100 is encouraging to improve U100 forecasts by downscaling information from Z500. However, improvements are only possible if the sufficient information about U100 can be regressed from Z500. 

\section{Regression from reanalysis input}\label{secRegression}
\begin{figure*}[t]
  \noindent
  \centering
  \includegraphics[width=39pc,angle=0]{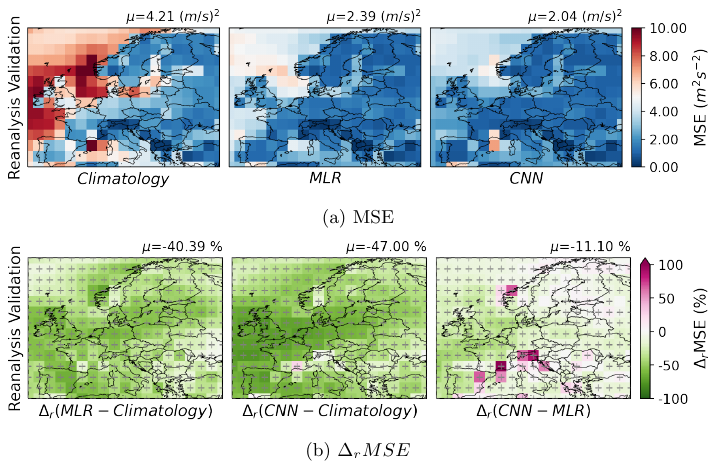}

  \caption{Maps of MSE (top) and $\Delta_r MSE$ (bottom) for the climatology, the MLR and the CNN estimated from validation reanalysis data. $\Delta_r MSE(model,bench)$ denotes the difference between the MSE of $model$ and the MSE of benchmark over the MSE of the latter. Thus, negative (positive) values correspond to improved (worsened) skill. 
  To assess the robustness of $\Delta_r MSE$, gray scatter points marked with "+" on the grid points denote statistically significant (0.01 significant level) improvements or degradations at those grid points. The spatial mean $\mu$ of the score and the $\Delta$ score are provided at the top of each subfigure.}
  \label{fig:RA_CL}
\end{figure*}

Whether more information from U100 can be regressed from Z500 using a non-linear model compared to a linear model is tested here by comparing U100 targets $Y_{test}$ to U100 regressed outputs $\hat{Y}_{test}$ from Z500 inputs  $\hat{X}_{test}$ on ERA5 reanalysis. Here, we employ rolling climatology as our benchmark. As mentioned before, this study defines rolling climatology as an interannual mean of the same calendar dates over past 15 years. We use the deterministic metric MSE to evaluate performance on the reanalysis test sets (figure \ref{fig:nestedCV}), with figure \ref{fig:RA_CL}a presenting the MSE maps for the climatology, the MLR, and the CNN across Europe. A lower MSE indicates a more accurate reconstruction of U100 reanalysis. As indicated at the top right corner of each subplot, the spatially averaged MSE across Europe for the climatology, the MLR, and the CNN are \replaced[id=auth]{4.21, 2.39, and 2.04}{4.34, 2.33, and 2.02} $(m/s)^2$, respectively. A conspicuous feature is the land-sea contrast, with higher MSE values at oceanic grid points and lower at terrestrial ones, due to higher wind speed and greater variability at sea. 

According to the definitions of the relative improvements in section \ref{secCase}\ref{subSecScores}, we display the relative improvements of MSE between models ($\Delta_r MSE(MLR,Climatology)$, $\Delta_r MSE(CNN,Climatology)$ and $\Delta_r MSE(CNN,MLR)$) in figure \ref{fig:RA_CL}b. It is evident that both the MLR and the CNN significantly outperform the climatology across all the grid points over Europe, with the spatially averaged $\Delta_r MSE(MLR,Cliamtology)$ and $\Delta_r MSE(CNN,Cliamtology)$ of \replaced[id=auth]{40.39\% and 47.00\%}{42.79\% and 49.44\%}, respectively. Especially in Western Europe, the MLR and the CNN exhibit approximately 50\% improvements. To explore the non-linearity in modeling the regression relationship between Z500 and U100 reanalysis, we also compared the $\Delta_r MSE(CNN,MLR)$ in figure \ref{fig:RA_CL}b, averaging around \replaced[id=auth]{11.10\%}{10.53\%} spatially, with the significant improvements observed across most grid points. 

The improvements of the CNN over the MLR (negative $\Delta_r MSE(CNN,MLR)$) demonstrate the presence of non-linear components in the Z500-U100 statistical relationship on the ERA5 reanalysis, which the MLR does not capture. These improvements manifest distinct spatial patterns; for instance, the degradation is seen over complex terrain like the Pyrenees, the Alps and the Scandinavian Mountains, whereas more substantial improvements occur over regions with greater wind variability, such as the North Sea and the Bay of Biscay. 

Overall, the CNN is more effective at extracting the information from Z500 reanalysis, thereby better representing the spatial variability of U100 over Europe. Together with the results of section \ref{secCase}\ref{subsecZ500SkilfulThan100uv}, these improvements motivate to use the non-linearity between Z500 and U100 reanalysis to regress U100 ensembles from Z500 ensembles.

\section{Improving ensemble forecasts using non-linear regression}\label{secForecasting}

In the previous section, we demonstrated that non-linearity can improve the reconstruction of U100 from Z500 reanalysis. In this section, we explore whether this non-linearity could be exploited to improve the sub-seasonal \replaced[id=rev1]{skill}{predictability} of U100 hindcasts compared to the linear model in figure \ref{fig:workflow}b. 
First, we apply the deterministic MLR and CNN as trained in the previous section to the 10-member dynamical Z500 hindcasts $X^M_{test}$, where $M=10$, from ECMWF to regress statistical ensembles following the workflow in figure \ref{fig:workflow}b. 
We evaluate the forecast\deleted[id=rev1]{ing} skill of these regressed ensembles $\hat{Y}^M_{test}$ from the MLR and the CNN on the sub-seasonal time scale, using the ECMWF hindcasts $Y^M_{test}$ as a benchmark. 
Subsequently, we assess the effectiveness of the stochastic perturbations $\widetilde{Y}^{M\times P}$, where $P=20$, following the same approach, when we need to attribute in the skill improvements to a better representation of the predictable part of the signal or to a better representation of the uncertainty. 

\subsection{Forecasts from deterministic regression}\label{subSecDirectForecasts}

\begin{figure*}[t]
  \noindent
  \centering
  \includegraphics[width=39pc,angle=0]{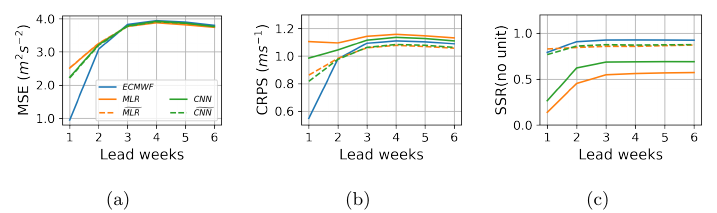}
  
  \caption{The spatial \replaced[id=auth]{mean}{median} of MSE, CRPS and SSR as a function of lead week across the Europe domain for U100 hindcasts from the ECMWF, the MLR, the CNN, the $\widetilde{MLR}$ and the $\widetilde{CNN}$.}
  \label{fig:curve_w1_w6}
\end{figure*}

As outlined in section \ref{secCase}, the ensemble uncertainty of Z500 hindcasts increases with lead time due to the chaotic nature of the atmosphere. In this section, we explore whether the non-linear CNN, compared to the MLR, can further improve U100 skill in the presence of hindcast uncertainty. In this subsection, ensembles from the ECMWF, the MLR, and the CNN each consist of 10 members, allowing us to conduct verifications with equal member size to avoid unfair comparison due to differing member size \citep{zamo2018estimation}.

Figure \ref{fig:curve_w1_w6} presents the spatial \replaced[id=auth]{mean}{median} of MSE, CRPS and SSR of the ECMWF, the MLR, the CNN and the perturbed version of the MLR ($\widetilde{MLR}$) and the CNN ($\widetilde{CNN}$), as functions of lead weeks. We first focus on the skill of the ECMWF, the MLR and the CNN. Only the medians derived from the bootstrap method (section \ref{secCase}\ref{subSecScores}) are provided to enhance the readability of these figures, while significant improvements in MSE and CRPS across models for different lead weeks are displayed in table \ref{table:direct_delta_w3_w6}. The MSE and the CRPS of these models increase with lead time and stabilize from the third week most likely due to increasing uncertainty. After lead week 4, there is a slight decrease in both MSE and CRPS for these models, likely because the hindcasts initialized in February include end dates in spring for lead week 5 and 6, which tend to be more predictable due to the lower U100 variability compared to winter dates \citep{cortesi2019characterization}. At lead weeks 1 and 2, the ECMWF exhibits skillful \deleted[id=rev1]{the} MSE and \deleted[id=rev1]{the} CRPS. At these lead times a significant fraction of the information from the initial conditions remains and the evolution of the state of the atmosphere is relatively well captured by the ECMWF IFS, making it difficult for the MLR and the CNN models, which only use Z500 hindcasts as inputs, to surpass. The CNN outperforms the MLR in terms of MSE \deleted[id=auth]{and CRPS }up to lead week 2, demonstrating that non-linearity can improve \replaced[id=auth]{the deterministic skill}{these skills} in the first lead weeks. \added[id=rev2]{However, this advantage diminishes as we progress into longer lead times. }Starting from lead week 3, the MSE curves for the MLR and the CNN converge towards the ECMWF benchmark. \added[id=rev2]{This convergence reflects a fundamental aspect of sub-seasonal forecasting, where the input-output relationships become increasingly dominated by unpredictable components at longer lead times. In such conditions, the complex non-linear mappings learned by CNN become less effective, and the simpler linear approximation provided by MLR appears to be equally capable of capturing the remaining predictable signals. }
Table \ref{table:direct_delta_w3_w6} further reveals that, at lead week 3, the $\Delta_r MSE(MLR,ECMWF)$ and $\Delta_r MSE(CNN,ECMWF)$ are marginal, remaining within a \replaced[id=auth]{1\%}{3\%}, but statistically significant. \added[id=rev2]{This observation suggests that increased model complexity alone cannot overcome the fundamental predictability barriers inherent in sub-seasonal forecasting.}

When analyzing the CRPS in figure \ref{fig:curve_w1_w6}b, the CNN exhibits a consistently lower CRPS throughout the sub-seasonal time scale compared to the MLR. However, both the MLR and the CNN still underperform relative to the ECMWF. Table \ref{table:direct_delta_w3_w6} demonstrates a significant improvement in CRPS for the CNN over the MLR by \replaced[id=auth]{2.43\%}{0.94\%}, suggesting that the non-linearity enhances the CRPS performance, yet the CNN remains \replaced[id=auth]{2.52\%}{3.50\%} worse than that of the ECMWF.

\begin{table}[t]
\caption{Spatially averaged $\Delta_r MSE$ (top) and $\Delta_r CRPS$ (bottom) comparing the improvements in the MLR over the ECMWF, the CNN over the ECMWF, the CNN over the MLR, and their perturbed version over the Europe domain for weeks 3 to 6. Negative values denote improved skill, while positive values signify a deterioration. 
Superscripts $^a$, $^b$, and $^c$ indicate statistical significance: $^a$ for $0 < p < 0.01$, $^b$ for $0.01 < p < 0.05$, $^c$ for $0.05 < p < 0.1$, where $p$ is the p-value estimated from the bootstrap method mentioned in section \protect\ref{secCase}\protect\ref{subSecScores}.
}\label{table:direct_delta_w3_w6}

\begin{center}
\footnotesize  
\begin{tabular}{cllll}



\hline\hline
$\Delta_r MSE$ (\%) & Week 3 & Week 4 & Week 5 & Week 6\\
\hline
($MLR$,$ECMWF$) & -0.84 $^a$ & -0.42 $^a$ & -0.6 $^a$ & +0.7 $^a$\\
($CNN$,$ECMWF$) & -0.05 $^a$ & +0.71 & +1.01 $^b$ & +1.65 $^a$\\
($CNN$,$MLR$) & +0.85 $^b$ & +1.17 $^a$ & +1.63 $^a$ & +0.92 $^b$\\
($\widetilde{MLR}$,$ECMWF$) & -0.56 $^a$ & -0.14 $^a$ & -0.33 $^a$ & +0.97 $^a$\\
($\widetilde{CNN}$,$ECMWF$) & -0.55 $^a$ & +0.2 $^a$ & +0.62 $^a$ & +1.52 $^a$\\
($\widetilde{CNN}$,$\widetilde{MLR}$) & +0.05 & +0.39 $^b$ & +1.0 $^a$ & +0.59 $^c$\\

\hline
$\Delta_r CRPS$ (\%) & Week 3 & Week 4 & Week 5 & Week 6\\
\hline
($MLR$,$ECMWF$) & +5.11 $^a$ & +4.81 $^a$ & +4.63 $^a$ & +4.97 $^a$\\
($CNN$,$ECMWF$) & +2.52 $^a$ & +2.77 $^a$ & +2.89 $^a$ & +3.07 $^a$\\
($CNN$,$MLR$) & -2.43 $^a$ & -1.92 $^a$ & -1.64 $^a$ & -1.8 $^a$\\
($\widetilde{MLR}$,$ECMWF$) & -2.87 $^a$ & -2.72 $^a$ & -2.75 $^a$ & -2.29 $^a$\\
($\widetilde{CNN}$,$ECMWF$) & -2.57 $^a$ & -2.19 $^a$ & -1.94 $^a$ & -1.68 $^a$\\
($\widetilde{CNN}$,$\widetilde{MLR}$) & +0.32 $^b$ & +0.56 $^a$ & +0.85 $^a$ & +0.64 $^a$\\
\hline

\end{tabular}
\end{center}
\end{table}

To understand why the CRPS of these statistical models is inferior to that of the ECMWF and to determine the contribution of non-linearity in improving CRPS, we employ SSR to assess ensemble reliability. Recall from section \ref{secCase}\ref{subSecScores} that an SSR value of 1 indicates a reliable ensemble, while values less than 1 suggest that the ensemble is under-dispersive, while values greater than 1 indicate an over-dispersive ensemble. 
In figure \ref{fig:curve_w1_w6}c, across the time scales of interest, the ensembles of the ECMWF, the MLR, and the CNN (shown in solid lines) are consistently under-dispersive and their dispersion increases but insufficiently after the first two weeks. The ECMWF approaches an SSR of 0.92, while the MLR and the CNN converge to 0.55 and \replaced[id=auth]{0.68}{0.59}, respectively, indicating these models lack spread, and so underestimate the forecast uncertainty. The low SSR of the MLR means its regressed members are close to each other, reflecting its ensemble mean. Notably, in the first two weeks, the dispersion of the MLR and the CNN rapidly increases, implying a swift growth in the uncertainty of the Z500 ensemble members used as inputs, with the CNN displaying greater dispersion than the MLR. Thus, even though non-linearity does not improve the MSE of ensemble mean, it contributes to improving the probabilistic metrics (SSR and CRPS) by better representing the regression model uncertainty.

\begin{figure*}[t]
  \noindent
  \centering
  \includegraphics[width=39pc,angle=0]{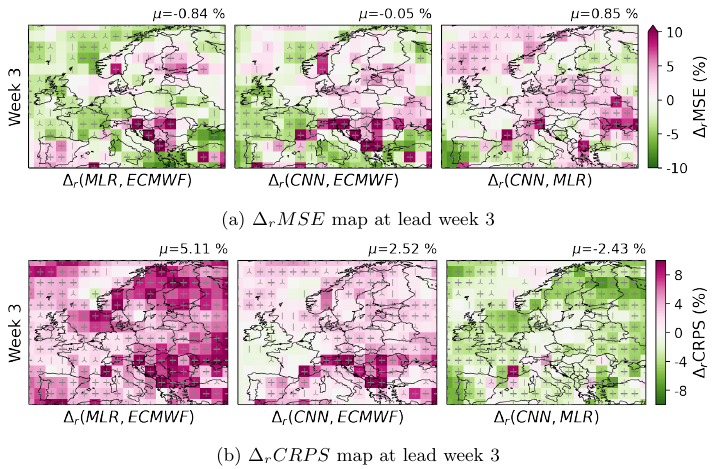}

  \caption{$\Delta_r MSE$ (top) and $\Delta_r CRPS$ (bottom) at lead week 3 comparing the improvements in the MLR over the ECMWF, the CNN over the ECMWF, and the CNN over the MLR. The significance of the results is represented in the same way as in figure \protect\ref{fig:RA_CL}. 
  }
  \label{fig:map_w3_deltaScore}
\end{figure*}

Aggregating these skills spatially obscures the spatial patterns, which are valuable for wind energy applications. 
In order to analyze the dependence of the skills on different grids, figure \ref{fig:map_w3_deltaScore} displays $\Delta_r MSE$ and $\Delta_r CRPS$ for lead week 3. Given the consistency in spatial patterns of the $\Delta_r MSE$ and $\Delta_r CRPS$ across various lead weeks, only the improvements for lead week 3 are presented. The maps detailing the $\Delta_r MSE$ and $\Delta_r CRPS$ from weeks 4 to 6 can be found in appendix (figures \ref{fig:map_w3_w6_delta_MSE} and \ref{fig:map_w3_w6_delta_CRPS}). For lead week 3, while the spatially averaged $\Delta_r MSE$ in table \ref{table:direct_delta_w3_w6} shows \replaced[id=auth]{marginal}{no significant} improvement for the MLR and the CNN relative to the ECMWF, distinct spatial patterns are observable in figure \ref{fig:map_w3_deltaScore}. The MLR and the CNN exhibit the more positive $\Delta_r MSE$ in some grids of Northern Europe than the ECMWF, indicating poorer performance in these regions. Conversely, in Western and Southeastern Europe, both the $\Delta_r MSE(MLR,ECMWF)$ and $\Delta_r MSE(CNN,ECMWF)$ are negative, with about a 5\% improvement in MSE noted in these grids relative to the ECMWF. Similar to the findings in section \ref{secRegression}, when applying the CNN to hindcasts, the CNN still struggles to effectively reconstruct wind speed in the grids with complex topography, resulting in poorer MSE and CRPS compared to the ECMWF and the MLR in these grids. The improvements of the CNN over the MLR also display a spatial distribution; the CNN performs better than the MLR in the North Sea, likely due to the greater wind speed variability in this region during winter and the CNN's non-linearity that more effectively reconstruct wind speed here (see section \ref{secRegression}). However, the CNN's MSE is approximately \replaced[id=auth]{4\%}{6\%} worse than the MLR's in Northern Europe.

For the maps of $\Delta_r CRPS$ in figure \ref{fig:map_w3_deltaScore}b, a spatial pattern similar to $\Delta_r MSE$ is observed, where regions with negative $\Delta_r MSE$ correspond to lower $\Delta_r CRPS$, even though most grid points across Europe underperform relative to the ECMWF. When considering $\Delta_r CRPS(CNN,MLR)$, the marginal but significant improvements (\replaced[id=auth]{2.43\%}{0.94\%}) are noted in most parts of Europe, except for Northern Europe and the grids with complex topography. This improvement is attributed to the CNN's ability to better fit the signal than the MLR, resulting in a higher variance of the regressed signal during model training. Therefore, when we use ensembles as inputs, the CNN can express more variance, thereby increasing ensemble dispersion.

\begin{figure*}[t]
  \noindent
  \centering
  \includegraphics[width=39pc,angle=0]{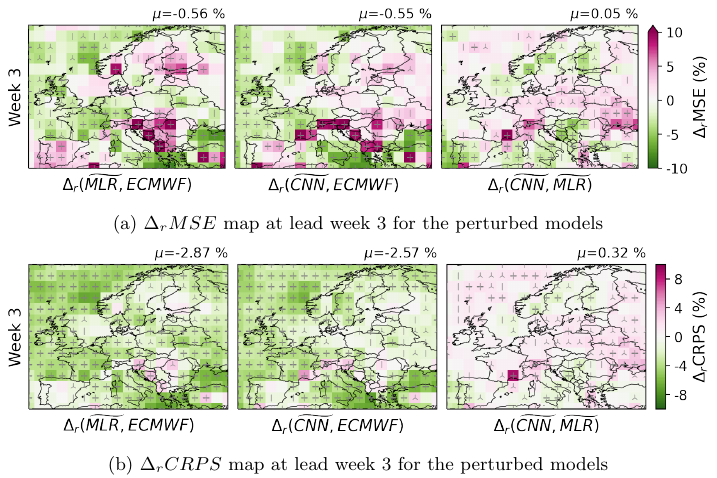}

  \caption{Same as figure \ref{fig:map_w3_deltaScore}. But comparing the improvements in the $\widetilde{MLR}$ over the ECMWF, the $\widetilde{CNN}$ over the ECMWF, and the $\widetilde{CNN}$ over the $\widetilde{MLR}$.}
  \label{fig:map_w3_deltaScore_stocha}
\end{figure*}

Overall, the non-linearity indeed contributes to the improvement of MSE and CRPS relative to the linear model during the first two lead weeks. However, the advantage of non-linearity in the deterministic skill measured by MSE diminishes after lead week 3. Despite this, in certain regions, such as Northwestern Europe, the non-linear model still outperforms the linear model. The CNN exhibits superior performance over the linear model in terms of CRPS by better representing the dispersion of the ensembles within the sub-seasonal timescale. Nevertheless, both the MLR and the CNN exhibit under-dispersion.

\subsection{Forecasts from stochastic perturbations}\label{subSecStochasticForecasts}

To address the issue of under-dispersion in statistical ensembles, we employ the method discussed in section \ref{secCase}\ref{subSecStochastic} to perturb the statistical ensembles of the MLR and the CNN. This subsection compares the perturbed versions of the MLR and the CNN, denoted as $\widetilde{MLR}$ and $\widetilde{CNN}$, respectively, which integrate model uncertainty during training, examining both the temporal evolution and spatial improvements of ensemble skills.

Dashed lines in figure \ref{fig:curve_w1_w6} display the temporal evaluation of forecast\deleted[id=rev1]{ing} skill of the perturbed ensembles $\widetilde{Y}^{M \times P}$ from the $\widetilde{MLR}$ and $\widetilde{CNN}$. 
In addition to the results for the non-perturbed models presented in section \ref{secForecasting}\ref{subSecDirectForecasts}, table \ref{table:direct_delta_w3_w6} also details the improvements at lead week 3 for these perturbed models. 
As expected, the perturbed ensembles from the $\widetilde{MLR}$ and the $\widetilde{CNN}$ do not exhibit great improvements in MSE compared to the MLR and the CNN, where improvements by these perturbed models over their direct models are less than 1\% and not statistically significant (not shown here),  which is not surprising considering that MSE is not a probabilistic score and does not derive benefits from the stochastic perturbations. However, the CRPS shows significant improvement from lead week 1 to 6. By lead week 3, the $\widetilde{MLR}$ and the $\widetilde{CNN}$ show over a \replaced[id=auth]{4\%}{6\%} improvement over their non-perturbed versions, represented by $\Delta_r CRPS(\widetilde{MLR}, MLR)$ and $\Delta_r CRPS(\widetilde{CNN}, CNN)$, which are not provided in table \ref{table:direct_delta_w3_w6}; however, they can be calculated using the table information such as 
$
\Delta_r CRPS (\widetilde{MLR}, MLR) = \Delta_r CRPS(\widetilde{MLR}, ECMWF) - \Delta_r CRPS(MLR, ECMWF) 
$  (same for the $\widetilde{CNN}$).
In the first two weeks, the $\widetilde{CNN}$ reports a lower CRPS than the $\widetilde{MLR}$, though the gap between the $\widetilde{MLR}$ and the $\widetilde{CNN}$ narrows after week 2, converging towards parity (the $\Delta_r CRPS(\widetilde{CNN},\widetilde{MLR})$ is less than 1\% in table \ref{table:direct_delta_w3_w6}). From week 3 onwards, the CRPS of the $\widetilde{MLR}$ and the $\widetilde{CNN}$ surpass that of the ECMWF, with an improvement of over \replaced[id=auth]{2\%}{3\%} compared to the ECMWF (in table table \ref{table:direct_delta_w3_w6}). The SSR (figure \ref{fig:curve_w1_w6}c) for the $\widetilde{MLR}$ and the $\widetilde{CNN}$ approaches \replaced[id=auth]{0.84}{0.85}, indicating that much of the under-dispersion has been corrected. Notably, the $\widetilde{MLR}$ and the $\widetilde{CNN}$, which incorporate the stochastic perturbations, exhibits greater reliability than the MLR and the CNN. This suggests that representing uncertainty from the training model is key to enhance the probabilistic skill of statistical ensembles.

The grid-level comparisons of $\Delta_r MSE$ and $\Delta_r CRPS$ for the $\widetilde{MLR}$ and the $\widetilde{CNN}$ are illustrated in figure \ref{fig:map_w3_deltaScore_stocha}, with a specific emphasis on week 3. Similarly, the maps for $\Delta_r MSE$ and $\Delta_r CRPS$ from week 4 to week 6 can be found in figures \ref{fig:map_w3_w6_delta_MSE}b and \ref{fig:map_w3_w6_delta_CRPS}b. The improvements in MSE for perturbed ensembles are temporally consistent, corroborating the findings presented in figure \ref{fig:map_w3_deltaScore_stocha}a. Across the European domain, \replaced[id=rev1]{except for the $\widetilde{MLR}$ in the North Sea}{ except for the North Sea of the $\widetilde{MLR}$}, there are improvements in CRPS for both models. The introduction of the perturbations reduces the $\Delta_r CRPS(CNN,MLR)$ of \replaced[id=auth]{2.43\%}{0.94\%} in figure \ref{fig:map_w3_deltaScore}b to the $\Delta_r CRPS(\widetilde{CNN},\widetilde{MLR})$ of \replaced[id=auth]{0.32\%}{0.28\%} in figure \ref{fig:map_w3_deltaScore_stocha}b, rendering them statistically insignificant over most grids. However, the slight improvement in CRPS for the $\widetilde{CNN}$ over the $\widetilde{MLR}$, although less than 1\%, is significant and predominantly observed in the North Sea and \replaced[id=auth]{Southern}{Eastern} Europe. \deleted[id=auth]{These improvements are \replaced[id=rev1]{consistent}{ consisted} with the grid points highlighted in figure \ref{fig:map_w3_deltaScore}b.}

Overall, the introduction of uncertainty quantification in these regression models has significantly improved the skills of the MLR and the CNN. Furthermore, for the perturbed ensembles, the spatially averaged improvements from the non-linearity are marginal.

\section{Conclusion}\label{secConclusion}

This study investigates the potential of non-linearity in improving the skill of U100 ensemble forecasts on the sub-seasonal timescale during European winter, while also exploring methods to better represent ensemble spread when applying regression models to sub-seasonal dynamical forecasts. Building on the works of \citet{alonzo2017modelling} and \citet{goutham2023statistical}, we employed both linear (MLR) and non-linear (CNN) models to compare their performance in modeling the regression relationship from Z500 reanalysis and hindcasts to surface variable U100.

Our results demonstrate that the CNN exhibits superior capability in capturing non-linear features compared to the MLR on the reanalysis dataset, thus enabling more accurate reconstruction of U100 over Europe, from knowledge of Z500 over the Europe-Atlantic domain, as indicated by the spatially averaged MSE. Specifically, the MSE for the MLR is \replaced[id=auth]{2.39}{2.33} $(m/s)^2$, while the CNN achieves an MSE of \replaced[id=auth]{2.04}{2.02} $(m/s)^2$. This represents approximately \replaced[id=auth]{11.10\%}{10.53\%} improvement in performance attributed to the non-linear characteristics on the reanalysis.

When applied to the ECMWF weekly Z500 hindcasts, the improvements of forecast\deleted[id=rev1]{ing} skill attributed to the non-linearity diminishes over time in terms of spatial averages. However, within specific regions (such as \replaced[id=auth]{North Sea}{Northwestern Europe}), the non-linearity still contributes to both MSE of ensemble mean and CRPS.

We also observed that the U100 ensembles generated by both the MLR and the CNN exhibited under-dispersion, representing an underestimation of forecast uncertainty, leading to poorer skills compared to the ECMWF dynamical ensembles. In other data driven weather forecast models\citep{robertson2015improving,bi2022pangu,kurth2023fourcastnet,chen2023fuxi,orth2014using}, it has also been found that the forecasts focused on the predictable part of the signal, and hence lacked dispersion. To address this issue, we introduced stochastic perturbations, which substantially improved the reliability and spread of the ensembles. As quantified by SSR, the MLR and the CNN exhibit the SSR values of 0.55 and \replaced[id=auth]{0.68}{0.59}, respectively. When the stochastic perturbations are introduced, the SSR increase\added[id=rev1]{s} to \replaced[id=auth]{0.84}{0.85} for both models.

Although the perturbed models ($\widetilde{MLR}$ and $\widetilde{CNN}$) still demonstrated some under-dispersion, their reliability improved relative to the original MLR and CNN. Notably, \replaced[id=rev1]{the dispersed ensemble had significantly enhanced probabilistic skill}{the dispersed ensemble spread significantly enhanced probabilistic skill}, with a corresponding improvement of approximately \replaced[id=auth]{4\%}{6\%} in CRPS for both the $\widetilde{MLR}$ and the $\widetilde{CNN}$ over their non-perturbed versions. Thus, better quantification of uncertainty on the sub-seasonal timescale is the key to improve sub-seasonal forecast\deleted[id=rev1]{ing} skill.

This study represents the first application of non-linear reconstruction of U100 from Z500 to improve sub-seasonal skill in U100 forecasts, as well as the introduction of the unexplained part of the signal by the regression model to improve ensemble reliability. Using historical data and only forecasting information of the large-scale part of the flow, as described by Z500, \replaced[id=rev1]{our approach combining statistical models with stochastic perturbations demonstrates improved probabilistic forecast skill over dynamical forecasts after lead week 3, particularly in CRPS, while maintaining comparable MSE.}{we have obtained forecast ensembles that perform better than the dynamical forecasts.} \replaced[id=auth]{While our research focuses on Z500-U100 relationships, this methodology has potential applications to other variables with regression relationships \citep{orth2014using, goutham2023statistical}.}{While our research focuses on the relationships between Z500 and U100, this methodology has potential applications to other variables connected through regression relationships, such as soil moisture and 2-meter temperature \citep{orth2014using}, and this methodology can be also applied to further improve forecasts with the ensemble hybrid method \citet{goutham2023statistical}.}

In terms of methodological choices, we utilized established MLR and CNN models. While more complex CNN architectures may yield further improvements, we found that designing a more realistic representation of regression uncertainty is crucial for significantly enhancing ensemble skill. Our approach did not account for spatial correlation among the stochastic perturbations, which presents an avenue for future research. Additionally, our calibration method (Mean Variance Adjustment, in appendix, section \ref{appendix:calibration}) faces consistency challenges across temporal and spatial dimensions, as it calibrates dynamical forecasts independently at each lead time and grid point, potentially disrupting the inherent spatial and temporal dependencies.

Future research directions include exploring whether the correlations in the regressed spatial patterns can be improved through a more sophisticated representation of stochastic perturbations. Additionally, it is worth investigating whether representing forecast uncertainty with ensemble spread could be improved by inputting the full ensemble distribution into the machine learning model, rather than applying the model separately to each ensemble member. Another direction for future research could examine whether learning directly from hindcasts/forecasts, rather than from reanalysis, and including the dynamical forecasts of target variables as inputs can help improve ensemble forecast\deleted[id=rev1]{ing} skill. Furthermore, explainable AI methods could help elucidate which input variables and which hidden units are actually responsible for skill improvements on average and geographically.

\clearpage
\acknowledgments
This work has been carried out at the Energy4Climate Interdisciplinary Center (E4C) of IP Paris and Ecole des Ponts ParisTech, which is in part supported by 3rd Programme d’Investissements d’Avenir [ANR-18-EUR-0006-02], and by the Foundation of Ecole polytechnique (Chaire “Défis Technologiques pour une Énergie Responsable” financed by TotalEnergies).

%
%
\datastatement
The ERA5 reanalysis data used in this study are available from the Climate Data Store (\url{https://cds.climate.copernicus.eu}). The ECMWF extended-range forecast data were obtained through Meteorological Archival and Retrieval System with institutional license. Researchers interested in sub-seasonal forecast and hindcast can access related datasets through the ECMWF S2S project (\url{https://www.ecmwf.int/en/research/projects/s2s}). All source codes have been made available on our GitHub repository (\url{https://github.com/TIANGANGLIN/s2s-wind-Non-linearity}). The repository also contains the trained parameters for both the MLR and CNN models, as well as their perturbed versions.

%






%



\appendix

\appendixtitle{a}

%



Table \ref{table:Notations} defines the symbols used in this article. For instance, $y_{g,n}$ represents the deterministic value of the input at the $t$-th of $T$ initialization times at a specific grid point, indicated by $g \in G$. Similarly, $y_{g,n, l}^m$ symbolizes the $m$-th member of a $M$-member ensemble at the $t$-th initialization time (of $T$ total times) for a given lead time $l$ at the corresponding grid point. For ensembles, the notation $\overline{\hat{y}^M} = \frac{1}{M} \sum_m^M y^m$ is used to denote the ensemble mean. The ensembles and the models that are perturbed are denoted with a tilde `` $\widetilde{}$ ".

\subsection{Pre-processing}\label{appendix:Preprocessing}

Our data pre-processing module involves sequential steps of spatial and temporal averaging and normalization. Reanalysis and ensembles are initially downsampled to a spatial resolution of 2.7 degrees (approximately 300 km) and a temporal resolution of 7 days. \added[id=auth]{Before normalization, we perform deseasonalization and detrending to isolate the underlying sub-seasonal signals. First, we remove the seasonal cycle by subtracting the mean climatology for each initialization date $t$. Unlike traditional grid wise deseasonalization, we calculate the spatial mean of climatology for each $t$ and subtract this seasonal pattern from all grid points to preserve spatial coherence in the anomalies. Second, we remove the long-term trend in the anomalies by fitting a linear model to the anomalies time series of spatially-averaged values, and then remove this estimated trend from all grid points, ensuring consistency in spatial patterns.} \deleted[id=auth]{Subsequently, we perform the normalization using a training dataset.}\replaced[id=auth]{Subsequently, the}{The} normalization of an input $x$ at the grid point $g$ is performed as $Normalization(x_{t,g}) = (x_{t,g} - \mu_g(X_{train}))/\sigma_g(X_{train})$, where $\mu_g(X_{train})$ and $\sigma_g(X_{train})$ respectively denote the mean and standard deviation along the initialized time dimension of the training dataset. \replaced[id=auth]{These}{The aforementioned} pre-processing operations are also applied independently to each forecast member $m$ and each lead week $l$ for forecasts and hindcasts.

\begin{table}[t]
\caption{Symbols in this study}\label{table:Notations}
\begin{center}
\begin{tabular}{ccc}
\hline\hline
Symbol    & Range     & Description           \\ 
\hline
x         &           & Input                 \\
y         &           & Target                \\
$\hat{y}$ &           & Output                \\
o         &           & Ground Truth          \\
$\widetilde{}$  &           & Perturbed version     \\ 
t         & 1, ..., T & Initialization time index      \\
l         & 1, ..., L & Lead time index           \\
g         & 1, ..., G & Grid point index      \\
m         & 1, ..., M & Ensemble-member index \\ 
\hline
\end{tabular}
\end{center}
\end{table}

\subsection{Scores}\label{appendix:score}
We employ the symbols defined in Table \ref{table:Notations} to calculate various verification scores, using the target variable $y$ as an example. The Mean Squared Error (MSE), the Continuous Ranked Probability Score (CRPS), the Spread Skill Ratio (SSR), the Mean Squared Skill Score (MSSS) and the Continuous Ranked Probability Skill Score (CRPSS) are initially calculated at the grid level to generate spatial distribution maps, then averaged spatially with cosine-latitude weights to illustrate the evolution of forecast\deleted[id=rev1]{ing} skill over lead weeks.

\subsubsection{Mean Squared Error (MSE)}
MSE is evaluated independently at each grid point for both the deterministic and probabilistic datasets. Specifically, for the regressed deterministic 100uv, represented as 
$\hat{y}$, MSE is computed as $MSE(y_{t,g}, \hat{y}_{t,g})$. For the regressed probabilistic 100uv ensembles $\hat{y}^M$, MSE is calculated using the ensemble mean $\overline{\hat{y}^M}$ at each grid point $g$ and for each lead week $l$, denoted as $MSE(y_{t,l,g}, \overline{y^M}_{t,l,g})$.

\begin{align}
MSE(y_{t,g}, \hat{y}_{t,g}) & = \frac{1}{T} \sum_t^T \left(y_{t,g} - \hat{y}_{t,g}\right)^2 \\
MSE(y_{t,l,g}, \overline{\hat{y}^M}_{t,l,g}) & = \frac{1}{T} \sum_t^T \left(y_{t,l,g} - \overline{\hat{y}^M}_{t,l,g}\right)^2 \\
\overline{\hat{y}^M} & = \frac{1}{M} \sum_m^M {\hat{y}^m}  
\end{align}



\subsubsection{Continuous Ranked Probability Score (CRPS)}

CRPS is used to assess the probabilistic skill of ensembles. Our benchmark dynamical hindcasts ensembles from ECMWF comprise merely 10 members, whereas our perturbed ensembles consist of 200 members. \citet{zamo2018estimation} and \citet{goutham2022skillful} indicate that an increase in ensemble member size improves the CRPS, making ensembles more skillful. To compare ensembles with varying member size, \citet{zamo2018estimation} suggests using a fair version of CRPS \citep{ferro2014fair} that estimates the CRPS as the member size approaches infinity. However, unlike the work of \citet{zamo2018estimation}, our perturbed ensembles do not satisfy the Fair CRPS assumption of exchangeability. This is because the perturbed ensembles are sampled based on each regressed member, which does not meet the requirements of being independently and identically distributed. Therefore, to ensure a fair comparison between the 10-member benchmark and the 200-member perturbed ensembles, following \citet{zamo2018estimation}'s approach, we downsample the 200-member perturbed ensembles to 10 members. To preserve the Cumulative Distribution Function (CDF) of the 200 members, we do not randomly select 10 members; instead, we use 10 quantiles of the 200 members as 10 ensemble members. Consequently, our comparison involves the 10-member ECMWF hindcasts and the 10 quantiles of the perturbed ensembles. \citet{zamo2018estimation} refers to the ensembles produced through this method as ``optimal ensembles," which achieve the minimal CRPS.

We employ a discrete version of CRPS \citep{zamo2018estimation}, calculating $CRPS_{l,g}$ for each lead time and each grid point.
\begin{equation}
\begin{split}
CRPS_{l,g} = & \frac{1}{t} \sum_{t}^{T} \left( \frac{1}{M} \sum_{m}^{M}|\hat{y}^m_{t,l,g}-y_{t,l,g}| \right. \\
& \left. -\frac{1}{2 M^2 } \sum_{m}^{M} \sum_{n}^{M}|\hat{y}^m_{t,l,g}-\hat{y}^n_{t,l,g}| \right)
\end{split}
\end{equation}


\subsubsection{Mean Squared Skill Score (MSSS) and Continuous Ranked Probability Skill Score (CRPSS)}\label{score_msss_crpss}
MSSS and CRPSS are used as relative scores. The subscript $ens$ denotes the ensembles of interest, while $ref$ refers to the reference with which comparisons are made. In this study, the 15-year rolling climatology is treated as reference.

\begin{align}
MSSS & = 1 - \frac{MSE_{ens}}{MSE_{ref}} \\
CRPSS & = 1 - \frac{CRPS_{ens}}{CRPS_{ref}}
\end{align}

\subsubsection{Spread Skill Ratio (SSR)}
SSR is the ratio of ensemble spread to the Root MSE of ensemble mean \citep{rasp2023weatherbench}, with ensemble spread represented by $Spread_{l,g}=\sqrt{\frac{1}{T} \sum_t^T var_m\left(y_{t, l, g}^m \right)^2}$, where $var_m$ is the variance across ensemble members. 

\begin{equation}
SSR_{l,g} = \frac{Spread_{l,g}}{\sqrt{MSE_{l,g}}}
\end{equation}

\subsubsection{Structural Similarity Index Measure (SSIM)}
SSIM comprises luminance $L$,  contrast $C$ and structure $S$ \citep{wang2004image}, with $\mu_g$, $\sigma_g$, and $cov_g$ denoting the spatial mean, the standard deviation, and the covariance, respectively.
\begin{align}
    SSIM_{l} & = L_{l} * C_{l} * S_{l} \\
    L_{l} & = \frac{1}{T*M} \sum_t^T \sum_m^M \frac{2 \mu_g (\hat{y}_{t, m, l}) \mu_g (y_{t, m, l})}{\mu_g^2 (\hat{y}_{t, m, l}) +\mu_g^2 (y_{t, m, l})} \\
    C_{l} & = \frac{1}{T*M} \sum_t^T \sum_m^M \frac{2 \sigma_g (\hat{y}_{t, m, l}) \sigma_g (y_{t, m, l})}{\sigma_g^2 (\hat{y}_{t, m, l})+\sigma_g^2 (y_{t, m, l})}  \\
    S_{l} & = \frac{1}{T*M} \sum_t^T \sum_m^M \frac{cov_g (\hat{y}_{t, m, l},y_{t, l})}{\sigma_g (\hat{y}_{t, m, l}) \sigma_g (y_{t, m, l})} 
    \label{eq:ssim}
\end{align}



\subsection{Calibration}\label{appendix:calibration}
we apply a simple and effective bias adjustment method, known as the Mean-Variance Adjustment (MVA). MVA is commonly utilized for the calibration of seasonal and sub-seasonal ensembles \citep{goutham2022skillful,goutham2023statistical,manzanas2019bias}. The mean and variance of ensembles are adjusted towards the mean and variance of reference data. We calibrate the forecasts and hindcasts using the same MVA as in \citet{goutham2022skillful}'s work.

\subsection{Models}\label{appendix:models}
\subsubsection{Multiple Linear Regression (MLR)}
\added[id=auth]{
This study employs a multiple linear regression (MLR) model to capture the linear relationship between Z500 and U100 in a spatially independent way. For each grid point $g$, a separate regression model is applied:$$\hat{Y}_{g} = \beta_{0,g} + \sum_{i=1}^{G} \beta_{i,g} \cdot X_{i} + \epsilon_{g}$$}

\added[id=auth]{
where $G$ denotes the total number of Z500 input grid points across the Europe-Atlantic domain, $\hat{Y}_{g}$ is the regressed U100 at grid point $g$, $\beta_{0,g}$ is the intercept term specific to grid point $g$, $\beta_{i,g}$ represents the regression coefficient linking input point $i$ to output point $g$, $X_{i}$ is the Z500 value at input grid point $i$, and $\epsilon_{g}$ is the error term.
}

\subsubsection{Convolutional Neural Network (CNN) - SmaAt-UNet}
\begin{figure*}[t]
 \noindent
  \centering
  \includegraphics[width=19pc,angle=0]{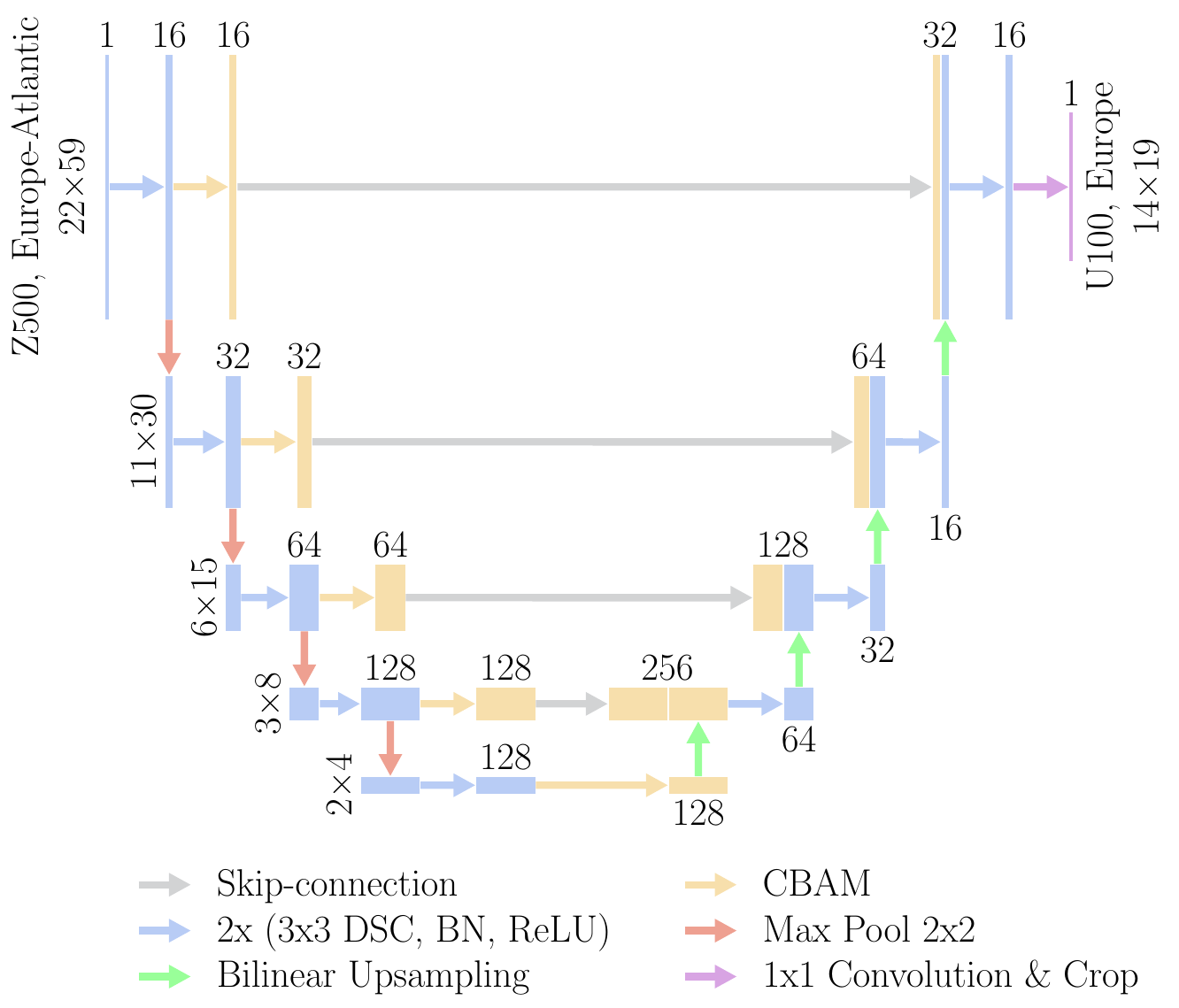}
  \caption{SmaAt-UNet architecture for regressing U100 from Z500. The diagram shows the network structure with Convolutional Block Attention Modules (CBAM), depthwise-separable convolutions (DSC), batch normalization (BN), and other key components that enable the model to capture the non-linear relationships between geopotential height and wind speed fields.}
  \label{fig:CNN}
\end{figure*}

\added[id=auth]{
The SmaAt-UNet in figure \ref{fig:CNN} is employed to capture the non-linear relationships between Z500 and U100. Our experiments showed that increasing model complexity through additional layers or more advanced architectures did not improve MSE performance on hindcasts, suggesting we had reached the complexity limit supported by our available data. The input is the Z500 across the European-Atlantic domain (22×59 grid), while the output is the regressed U100 over the Europe domain (14×19 grid). The encoder of SmaAt-UNet includes four downsampling stages, each comprising: (1) two consecutive 3×3 depthwise-separable convolutions (DSC), followed by batch normalization (BN) and ReLU activation; (2) a 2×2 max-pooling layer to reduce spatial dimensions; (3) Convolutional Block Attention Modules (CBAM) are placed after the first double convolution and after each encoder to amplify important features. As the network deepens, the number of feature channels increases (from 16 to 128), while the spatial resolution decreases. The decoder also comprises four stages, reconstructing U100 through bilinear interpolation. Key characteristics at each layer include: (1) skip connections from corresponding encoder layers to preserve detailed information; (2) two consecutive depthwise-separable convolution operations identical to those in the encoder; (3) bilinear upsampling operations to double the spatial resolution of feature maps. Ultimately, a 1×1 convolution operation is applied to map the feature space and a crop operation is applied to truncate it to the target resolution of a 14×19 grid for U100. }

\added[id=auth]{Both MLR and CNN are implemented on Pytorch \citep{paszke2019pytorch} and trained using Adam optimizer to minimize MSE, with hyperparameters (learning rate and weight decay) optimized via Optuna \citep{akiba2019optuna}.}

\subsection{Forecast\deleted[id=rev1]{ing} skills of z500 and 100uv}\label{appendix:hypothese}
\begin{figure*}[t]
 \noindent
  \centering
  \includegraphics[width=39pc,angle=0]{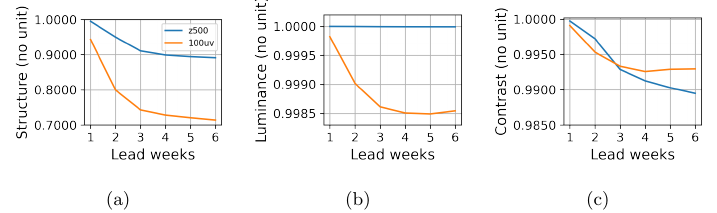}

  \caption{Same as figure \protect\ref{fig:check_hypo_1}, but for Structure, Luminance and Contrast.}
  \label{fig:check_hypo_appendix}
\end{figure*}

The three components of SSIM, calculated using the formula presented in appendix, section \ref{appendix:score}, are displayed in figure \ref{fig:check_hypo_appendix}. Luminance and Contrast exhibit small variation across the lead weeks, indicating that the spatial mean and the variance of ensemble members remain relatively stable over time. As illustrated in figure \ref{fig:check_hypo_1}c, the decrease in SSIM across the lead weeks primarily stems from a decrease in Structure. This suggests significant variations in the covariance between ensemble members and the corresponding reanalysis, inaccurately representing the spatial patterns of the atmospheric states.

\subsection{Skill improvements of the 100uv ensembles from the ECMWF, the MLR, the CNN, the $\widetilde{MLR}$, the $\widetilde{CNN}$ from week 4 to week 6}\label{appendix:deltaSkillW4_W6}

\begin{figure*}[t]
  \noindent
  \centering
  \includegraphics[width=30pc,angle=0]{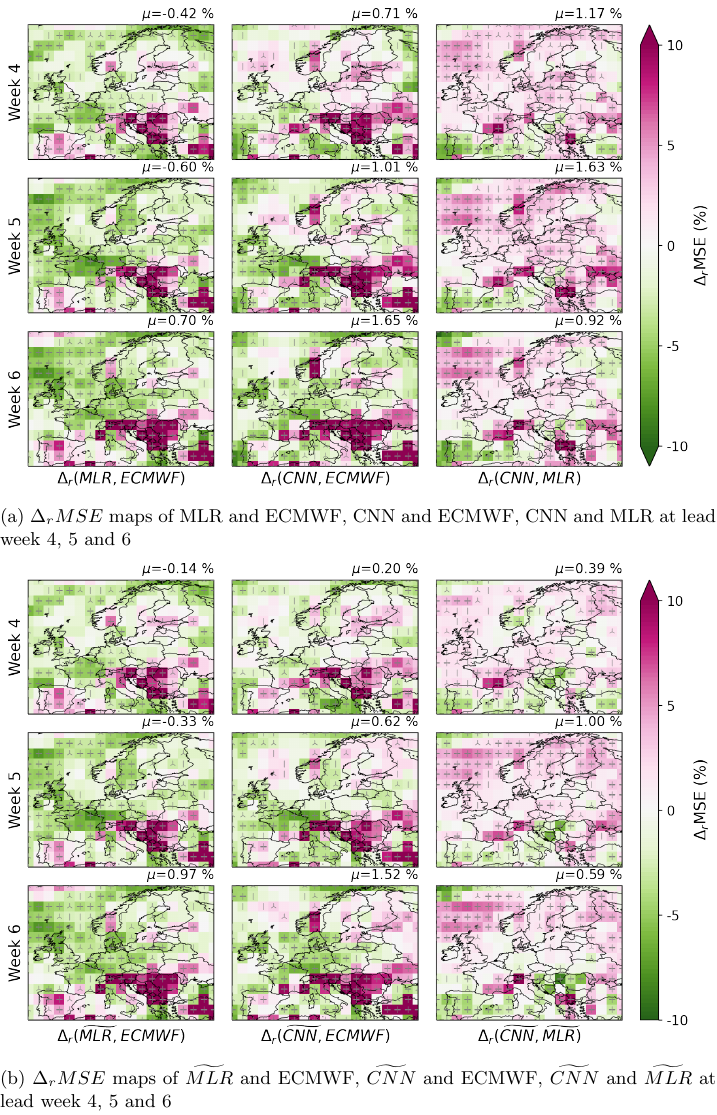}

  \caption{Same as figures \protect\ref{fig:map_w3_deltaScore} and \protect\ref{fig:map_w3_deltaScore_stocha}, but for the $\Delta_r MSE$ \replaced[id=auth]{from week 4 to week 6}{from week 4 to week 6}}
  \label{fig:map_w3_w6_delta_MSE}
\end{figure*}

\begin{figure*}[t]
  \noindent
  \centering
  \includegraphics[width=30pc,angle=0]{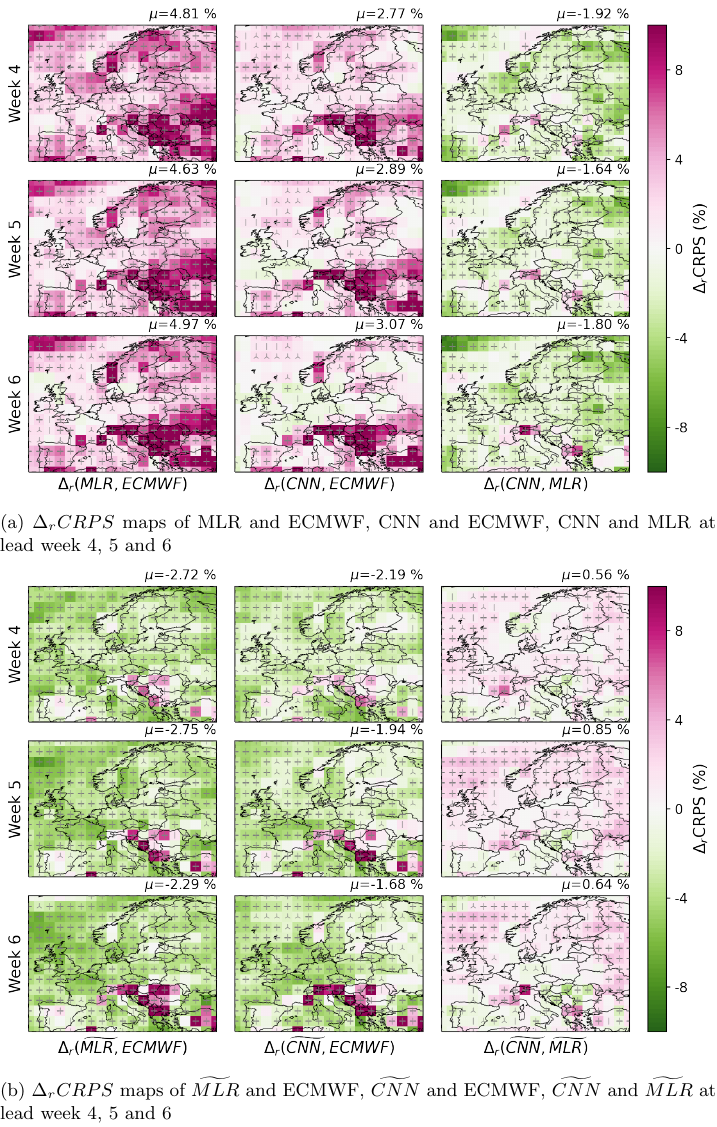}

  \caption{Same as figure \protect\ref{fig:map_w3_deltaScore} and \protect\ref{fig:map_w3_deltaScore_stocha}, but for the $\Delta_r CRPS$ 
  \replaced[id=auth]{from week 4 to week 6}{from week 4 to week 6}}
  \label{fig:map_w3_w6_delta_CRPS}
\end{figure*}

In the main text, we presented the $\Delta_r MSE$ and $\Delta_r CRPS$ for the models during lead week 3. Here, we extend our analysis from lead week 4 to lead week 6, demonstrating $\Delta_r MSE$ and $\Delta_r CRPS$ for this period in figures \ref{fig:map_w3_w6_delta_MSE} and \ref{fig:map_w3_w6_delta_CRPS}. Notably, during lead weeks 4 and 5, from the $\Delta_r MSE(MLR,ECMWF)$ and the $\Delta_r MSE(CNN,ECMWF)$ both the MLR and the CNN exhibit significant improvements over the ECMWF primarily in France, Spain and Eastern Europe; however, by lead week 6, a notable degradation is observed in the North Atlantic. The comparison between the $\Delta_r MSE(MLR,ECMWF)$ and the $\Delta_r MSE(CNN,ECMWF)$ highlights these trends. The improvements attributed to non-linearity are primarily concentrated in the North Sea, Germany, and Eastern Europe. However, the CNN exhibits worse MSE compared to the MLR in regions with complex terrain and Southern Europe. Similarly, the perturbed versions, $\Delta_r MSE(\widetilde{CNN},\widetilde{MLR})$ in figure \ref{fig:map_w3_w6_delta_MSE}b demonstrates similar spatial distribution patterns as $\Delta_r MSE(CNN,MLR)$.

Figure \ref{fig:map_w3_w6_delta_CRPS} illustrate the $\Delta_r CRPS$ from lead week 4 to lead week 6, where the MLR and the CNN demonstrate worse CRPS than the ECMWF. However, the benefits of non-linearity, indicated by more negative values in CRPS as analyzed in the main text, are evident. For the $\widetilde{MLR}$ and the $\widetilde{CNN}$, the introduction of the stochastic perturbations consistently aids CRPS across the sub-seasonal timescales, although the non-linear assistance diminishes with the increasing lead weeks.

\subsection{Spatially averaged skills of the 100uv ensembles from the ECMWF, the MLR, the CNN, the $\widetilde{MLR}$, the $\widetilde{CNN}$ on forecasts}\label{appendix:res_fc}
\begin{figure*}[t]
  \noindent
  \centering
  \includegraphics[width=39pc,angle=0]{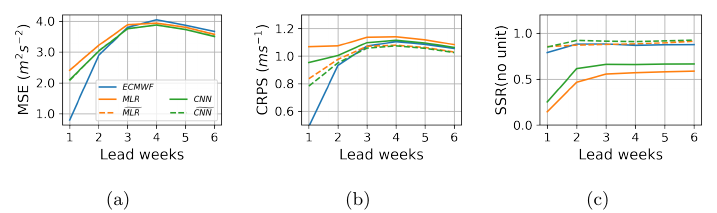}
  \caption{The same as figure \protect\ref{fig:curve_w1_w6}, but on forecasts}
  \label{fig:curve_w1_w6_fc}
\end{figure*}

Similar to figure \ref{fig:curve_w1_w6}, we present the spatially averaged MSE, CRPS, and SSR for the ECMWF, the MLR, the CNN, the $\widetilde{MLR}$ and the $\widetilde{CNN}$ across the forecasts dataset in figure \ref{fig:curve_w1_w6_fc}. It is noteworthy that the size of the forecasts dataset is considerably smaller than that of the hindcasts, leading to less representative results from the forecasts. Nonetheless, the conclusions drawn from the forecasts are consistent with those from the hindcasts dataset.

Compared to figure \ref{fig:curve_w1_w6}, the performance of the MLR, the CNN, the $\widetilde{MLR}$ and the $\widetilde{CNN}$ on forecasts aligns with their performance on hindcasts. For MSE, within the initial two weeks, the ECMWF benefits from its complex dynamics, exhibiting better MSE of the ensemble mean. From lead week 2, the MLR and the CNN demonstrate better MSE than the ECMWF, and the $\widetilde{MLR}$ and the $\widetilde{CNN}$ do not further improve the MSE of the MLR and the CNN. Regarding CRPS, the non-linear CNN maintains a lower CRPS in the first four weeks, underscoring the benefits of non-linearity in the enhancing probabilistic forecast\deleted[id=rev1]{ing} skill. When the stochastic perturbations are introduced, the impact of non-linearity on CRPS diminishes with increasing lead weeks. As for SSR, the MLR and the CNN's SSR are under-dispersive. The addition of the stochastic perturbations significantly improve the reliability of forecasts, but insufficiently.

\bibliographystyle{ametsocV6}
\bibliography{references}

\end{document}